\documentclass{article}
\usepackage{ijcai24}

\usepackage{times}
\usepackage{soul}
\usepackage{url}
\usepackage{hyperref}
\hypersetup{hidelinks}
\usepackage[utf8]{inputenc}
\usepackage[small]{caption}
\usepackage{graphicx}
\usepackage{amsmath}
\usepackage{amsthm}
\usepackage{booktabs}
\usepackage{algorithm}

\usepackage[switch]{lineno}
\usepackage{dblfloatfix}
\usepackage{multirow}
\usepackage{subcaption} 
\usepackage[noend]{algpseudocode}

\title{One for All: A General Framework of LLMs-based Multi-Criteria Decision Making on Human Expert Level}

\author{
Hui Wang$^{1,2}$
\and
Fafa Zhang$^{1,2}$
\and
Chaoxu Mu$^{1,2}$\\
\affiliations
$^1$Anhui University\\
$^2$Anhui Provincial Key Laboratory of Security Artificial Intelligence\\
\emails
h.wang.13@ahu.edu.cn,
wa23301160@stu.ahu.edu.cn,
cxmu@tju.edu.cn
}
\begin{document}

\maketitle

\begin{abstract}
Multi-Criteria Decision Making~(MCDM) is widely applied in various fields, using quantitative and qualitative analyses of multiple levels and attributes to support decision makers in making scientific and rational decisions in complex scenarios. However, traditional MCDM methods face bottlenecks in high-dimensional problems. Given the fact that Large Language Models~(LLMs) achieve impressive performance in various complex tasks, but limited work evaluates LLMs in specific MCDM problems with the help of human domain experts, we further explore the capability of LLMs by proposing an LLM-based evaluation framework to automatically deal with general complex MCDM problems. Within the framework, we assess the performance of various typical open-source models, as well as commercial models such as Claude and ChatGPT, on 3 important applications, these models can only achieve around 60\% accuracy rate compared to the evaluation ground truth. Upon incorporation of Chain-of-Thought or few-shot prompting, the accuracy rates rise to around 70\%, and highly depend on the model. In order to further improve the performance, a LoRA-based fine-tuning technique is employed. The experimental results show that the accuracy rates for different applications improve significantly to around 95\%, and the performance difference is trivial between different models, indicating that LoRA-based fine-tuned LLMs exhibit significant and stable advantages in addressing MCDM tasks and can provide human-expert-level solutions to a wide range of MCDM challenges. 
\end{abstract}

\begin{figure*}[htb]
    \centering
       \includegraphics[width=\textwidth]{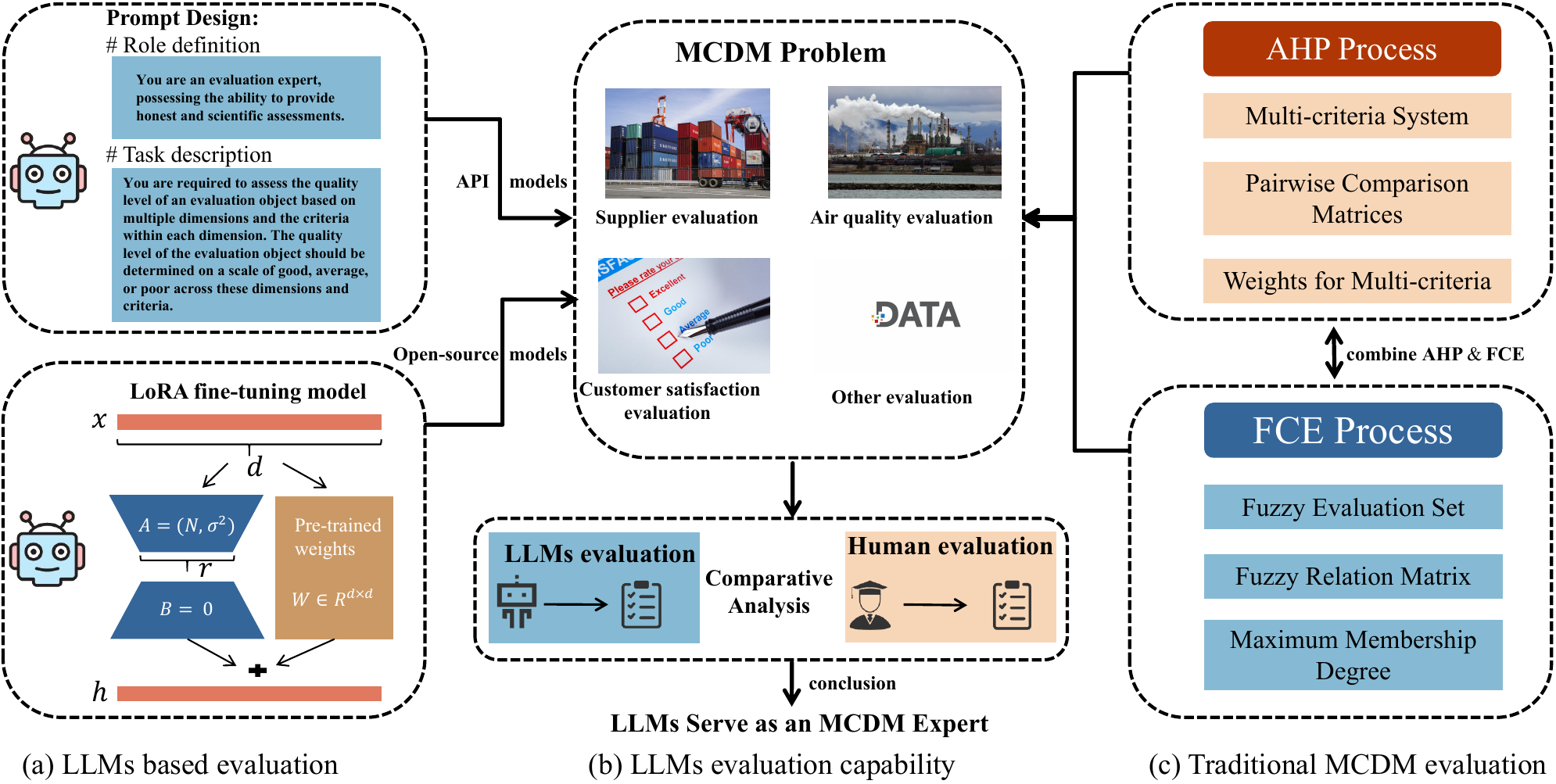}
    \caption{Overview of the proposed LLMs-based MCDM Framework. (a) The API models and the LoRA fine-tuned models are evaluated separately for MCDM. (b) Comparative analysis of the two sets of evaluation results. (c) Conducting MCDM evaluation combining traditional Models.}
    \label{fig:fig1}
\end{figure*}

\section{Introduction}

Multi-Criteria Decision Making~(MCDM) is designed to address complex decision-making problems that involve multiple criteria. Quite a lot of work indicates that MCDM techniques can help decision makers select the optimal option among many alternatives by comprehensively evaluating multiple criteria. Therefore, MCDM becomes an invaluable tool in a wide range of fields. Over the years, numerous techniques have been developed to address MCDM problems. It is particularly effective in applications such as investment analysis, project evaluation, economic efficiency evaluations, and employee performance evaluations~\cite{1-aruldoss2013survey,2-triantaphyllou2000multi,3-taherdoost2023multi}.

Traditional methods in MCDM include the Analytic Hierarchy Process (AHP)~\cite{4-belay2022ahp,5-aliyev2020fuzzy}, which decomposes a problem into multiple hierarchical levels and constructs a judgment matrix to calculate the weights of elements at each level. Another widely used method is the Technique for Order Preference by Similarity to Ideal Solution (TOPSIS)~\cite{6-trung2021multi,7-bali2023commercial}, which evaluates alternatives by calculating their geometric distances from the ideal and negative ideal solutions. In addition, the VIKOR method constructs a loss function to compute a comprehensive score for each alternative, selecting the one closest to the ideal solution. This approach is particularly effective in addressing decision-makers' preferences and trade-offs~\cite{9-vaid2022application}. The Fuzzy Comprehensive Evaluation~(FCE) method~\cite{8-masdari2021service}, rooted in fuzzy mathematics, is well suited to handle uncertainty and vagueness in decision making. Integrates the results of the evaluation using fuzzy matrices to process and interpret fuzzy information effectively. In addition, several studies have investigated the combination of AHP and FCE for MCDM~\cite{10-abdullah2023multi,11-afolayan2020performance,12-xu2023multi,13-ecer2022multi}, utilizing their complementary strengths. These methods have been extensively applied in fields such as project evaluation, investment decision making, resource allocation, and supply chain evaluation. Using these tools, decision makers are empowered to make more informed and rational decisions in complex and uncertain environments.

With the rapid development of Large Language Models~(LLMs), groundbreaking advancements have been achieved in various technological domains. Models such as ChatGPT and LLaMA, using their advanced natural language processing capabilities, are redefining human-machine interaction, automated content generation, and decision support at multiple levels~\cite{15-wang2023chatgpt,16-zhao2024llama}. In traditional MCDM, decision-making processes heavily rely on human experts or decision makers to assess problems and evaluate solutions. These methods involve quantifying criteria, comparing alternatives, analyzing trade-offs, and ultimately making final selections. Such processes often require extensive data collection, expert judgment, and a comprehensive evaluation of multiple criteria. However, the emergence of LLMs is revolutionizing the operational approach of MCDM methods. With their powerful computational and learning capabilities, LLMs can process and analyze large amounts of textual data in extremely short time frames. This allows them to provide more precise, efficient, and comprehensive information support for decision-making processes~\cite{14-wang2024can}.

LLMs such as LLaMA, Qwen, and ChatGLM demonstrate strong performance in natural language processing tasks~\cite{17-bai2023qwen,18-glm2024chatglm,19-yang2023baichuan}. However, directly applying these pre-trained models to specific domains may not yield optimal results. To address this limitation, the performance of LLMs in specialized domains can be significantly improved by fine-tuning~\cite{20-hu2021lora}. Fine-tuning introduces adaptive adjustments that effectively capture domain-specific information, thereby improving the accuracy and applicability of the models. Recent studies~\cite{21-wang2023huatuo,22-zheng2024fine,23-han2024parameter} have demonstrated that fine-tuning methods leveraging domain-specific knowledge can substantially enhance model capabilities and performance in targeted applications. However, there is still no single general MCDM framework that can be used as a human expert to deal with different MCDM tasks automatically.

Therefore, in this paper, we propose a general MCDM framework that can serve as human experts on different MCDM applications. To validate the efficiency of the proposed framework, we tested three applications: supply chain evaluation, air quality evaluation, and customer satisfaction evaluation. The main contributions of this paper are summarized as follows:
\begin{enumerate}
    \item We propose and implement a human-expert-level LLM-based MCDM assessment framework that can evaluate multiple decision-making scenarios more efficiently by utilizing advanced natural language processing techniques to improve the automation and intelligence of handling MCDM tasks.
    \item The performance of several open-source LLMs, such as Qwen2, ChatGLM4, and Llama3, as well as commercial models such as Claude and ChatGPT, has been systematically evaluated. The experimental results demonstrate that LLMs do not show good performance in processing MCDM scenarios, indicating the requirements of combining other techniques to improve the ability.
    \item To improve the performance of the models in MCDM tasks, we employ the LoRA fine-tuning technique. With a small amount of data for tuning, the models achieved notable improvements across multiple MCDM datasets, significantly reducing the performance gap between different LLMs.
\end{enumerate}

\section{Related Work}
Traditional MCDM methods have been widely applied in areas such as supplier evaluation and environmental impact assessment. These methods typically rely on the judgments of human experts involving the weighing and comparison of multiple criteria. Although effective in practical applications, traditional MCDM methods can be time-consuming and resource-intensive, particularly when dealing with large datasets and complex scenarios, due to their heavy reliance on manual input. In recent years, LLMs have emerged as a promising solution to address these challenges. Several studies have explored the use of LLMs as virtual judges to automate the assessment and decision-making process~\cite{30-li2024generation,33-pasch2025llm,34-bennie2025panda}. For example, Wang and Wu proposed the application of ChatGPT for supplier evaluation, demonstrating that its performance was highly consistent with the results of manual expert evaluations~\cite{14-wang2024can}. Similarly, Svoboda et al. introduced a multi-criteria decision analysis framework for cybersecurity that combines AHP with GPT-4~\cite{31-svoboda2024enhancing}. This framework automates decision making while improving both the efficiency and reliability of decisions by using GPT-4 as a virtual expert. Furthermore, Dong et al.~\cite{32-dong2024can} introduced linguistic uncertainty estimation to improve evaluation consistency for high-confidence samples. Their findings showed that in certain tasks, the performance of LLM-based evaluations matched or even exceeded that of human experts. These studies highlight a more reliable and scalable approach to leveraging LLMs for personalized evaluations, offering innovative opportunities to automate and enhance decision-making processes. However, a single general MCDM framework should be explored to serve as a general expert in solving different MCDM tasks to reduce the dependence on domain knowledge.

\section{METHODOLOGY}
In this study, we compare the traditional AHP-FCE method with LLMs to explore their applications in MCDM problems. Three data sets were used for the experiments: a supplier evaluation dataset, a customer satisfaction assessment dataset, and an air quality evaluation data set. Each data set contains multiple dimensions for evaluation. The primary objective of this study is to evaluate the importance and weights of these dimensions using data sets and to compare the results with those obtained from traditional AHP-FCE methods. The comparison aims to verify whether LLMs can effectively assist or even replace traditional MCDM methods in practice.



AHP is a decision-making method that decomposes elements related to decision making into multiple levels, such as objectives, criteria, and alternatives, and performs both qualitative and quantitative analysis. Depending on the nature of the problem and the ultimate goal, AHP breaks the problem down into various factors. By analyzing the interrelationships and dependencies among these factors, it aggregates and synthesizes them at different levels to create a multilevel analytical structure. Ultimately, the problem is solved by determining the relative importance weights of the lowest-level elements in relation to the highest-level objectives. FCE is a comprehensive evaluation method based on fuzzy mathematics, using the affiliation theory of fuzzy sets to transform qualitative evaluations into quantitative ones. Conducting an overall evaluation of objects or phenomena constrained by multiple factors, FCE is particularly effective in dealing with fuzzy and difficult-to-quantify problems, providing clear and systematic results, and integrating both qualitative and quantitative factors, expanding the amount of information available, and enhancing the reliability of evaluation outcomes. Consequently, FCE is widely applied in state evaluation.
In summary, AHP and FCE each have distinct advantages and are commonly employed in complex decision-making and comprehensive evaluation processes to ensure scientific and rational decision-making. The combination of AHP and FCE synthesizes both qualitative and quantitative analysis, improving the accuracy of weight determination and enhancing the model's adaptability. This integrated approach results in more scientific, comprehensive, and reliable decision-making and evaluation. 

\begin{algorithm}[!b]
\caption{LLM-based MCDM Evaluation}
\label{alg:dual_path}
\begin{algorithmic}[1]
\Require  Pretrained LLMs $\mathcal{M} = \{M_{\theta_1},...,M_{\theta_k}\}$
\Require  API LLMs  $\mathcal{X} = \{X_{\theta_1},...,X_{\theta_k}\}$
\Require MCDM problem set $P$, system prompts $\mathcal{S}$
\Require LoRA hyperparameters: rank $r$, training steps $T$
  
\State Initialize result repository $R \gets \emptyset$

\For{each problem $p \in P$} 

  \For{each prompt template $S \in \mathcal{S}$} 
    \For{each model $X_\theta \in \mathcal{X}$}
      \State Construct context $C \gets S \oplus p$
      \State $\alpha_{\text{vanilla}} \gets M_\theta(C)$
      \State Store $(p, \alpha_{\text{vanilla}}, S, X_\theta)$ in $R$
    \EndFor
  \EndFor
  \For{each model $M_\theta \in \mathcal{M}$}
    \State Initialize adapters: $A \sim \mathcal{N}(0,\sigma^2), B \gets 0$
    \For{$t \gets 1$ \textbf{to} $T$}
      \State $(\nabla_A, \nabla_B) \gets \frac{\partial \mathcal{L}}{\partial (A,B)}(p, M_\theta)$
      \State Update: $A \gets A - \eta\nabla_A,\ B \gets B - \eta\nabla_B$
    \EndFor
    \State $\tilde{M}_\theta \gets M_\theta + \frac{\alpha}{r}AB^\top$ 
    \For{each prompt template $S \in \mathcal{S}$}
      \State Construct context $C \gets S \oplus p$
      \State $\alpha_{\text{tuned}} \gets \tilde{M}_\theta(C)$
      \State Store $(p, \alpha_{\text{tuned}}, S, M_\theta)$ in $R$
    \EndFor
  \EndFor
\EndFor
\Return $R$
\end{algorithmic}
\end{algorithm}

The AHP-FCE based method consists of the following steps.
\begin{enumerate}
    \item Clarifying the indicators for the object to be evaluated and determining the set of factors for the object to be evaluated.
    
    \item Determine the set of weights \( W_i \) using the AHP method.
  \begin{equation}\label{eq:weights}
W_i = (w_1, w_2, \dots, w_n)
\end{equation}

    \item Create a collection of rubrics.
    \begin{equation}\label{eq:rubrics}
    V_i = \{v_1, v_2, \dots, v_w\}
    \end{equation}
    
    \item Create a fuzzy composite rating matrix \( R \).
     \begin{equation}\label{eq:matrix_R}
    R =
    \begin{bmatrix}
    R_1 \\
    R_2 \\
    \vdots \\
    R_m
    \end{bmatrix}
    =
    \begin{bmatrix}
    r_{11} & r_{12} & \dots & r_{1n} \\
    r_{21} & r_{22} & \dots & r_{2n} \\
    \vdots & \vdots & \ddots & \vdots \\
    r_{m1} & r_{m2} & \dots & r_{mn}
    \end{bmatrix}
    \end{equation}
    In equation~\ref{eq:matrix_R}, \( R_i = (r_{i1}, r_{i2}, \dots, r_{in}) \) reflects the independent evaluation result of the factor \( i \).
    
    \item Synthesize the vector of FCE results. The appropriate operator is selected to synthesize the weight set and the FCE matrix to produce a vector of FCE for each evaluation factor.
    
    \item Calculate the complete score of the indicator system using the weighted average method to derive the FCE conclusion.
\end{enumerate}

Furthermore, we propose the algorithm which aims to build a flexible and scalable automated decision support system by integrating MCDM problems with generative LLMs. The core idea is to leverage multiple pre-trained generative language models and diverse system prompts to process a given set of MCDM problems, generate detailed solutions, and store the questions and answers in a structured format for further analysis and evaluation. The pseudo code is given as Algorithm~\ref{alg:dual_path}.

\subsection{Evaluation with LLMs}
As shown in Fig.~\ref{fig:fig1}, this paper presents a framework for the integration of LLM and MCDM. The framework employs two approaches during the model preparation phase: one involves using direct prompts through APIs for design, while the other utilizes open-source models that are fine-tuned on small datasets. The language models prepared through these two methods are then evaluated on the MCDM dataset to generate results. In the final step, the effectiveness of language models in automating and improving decision-making accuracy is validated by comparing the evaluation results of the LLMs with those obtained from expert evaluations. We evaluated three datasets using the LLMs, focusing on its application to MCDM tasks. Three different approaches are employed: zero-shot, few-shot, chain of thought. Furthermore, for pre-trained open source models, we employed LoRA fine-tuning to improve the performance. 

\subsubsection{Zero-shot and few-shot prompts}
Zero-shot and few-shot are two common task settings to evaluate a model's ability to reason and make decisions under varying amounts of information~\cite{28-liu2024logprompt}. Zero-shot refers to a scenario where the task description is provided directly without any examples, and the model is required to perform reasoning and decision analysis based solely on the given description. In contrast, few-shot involves providing a small number of task-specific examples to help the model understand the task's requirements and decision criteria. These examples serve as guidance, allowing the model to better understand how to perform the task and improving its performance in MCDM tasks. In the context of MCDM, zero-shot and few-shot provide distinct approaches to solving decision-making problems. The zero-shot setting evaluates the model's generalization ability, while the few-shot setting allows the model to better comprehend and address complex decision-making tasks with the help of a small number of examples.

\begin{table}[!b]
    \centering
    \begin{tabular}{lcccc}
        \hline
       Dimension & Dimension  & Criteria & Criteria  & Overall  \\
       &Weight&&Weight&Weight\\
        \hline
        & & Advanced & 47.29\% & 12.40\% \\
        Delivery  & 26.22\% & On time & 16.99\% & 4.45\% \\
        Status& & Late & 28.44\% & 7.46\% \\
        & & Canceled & 7.29\% & 1.91\% \\
        \hline
        & & $<$ -1 & 3.33\% & 1.88\% \\
        & & -1 to -0.5 & 6.34\% & 3.58\% \\
        Profit  & 56.50\% & -0.5 to 0 & 12.90\% & 7.29\% \\
        Margin& & 0 to 0.5 & 26.15\% & 14.77\% \\
        & & $>$ 1 & 51.28\% & 28.97\% \\
        \hline
        & & Standard & 56.50\% & 6.64\% \\
        Shipping  & 11.75\% & Second & 26.22\% & 3.08\% \\
       Mode & & First & 11.75\% & 1.38\% \\
        & & Same Day & 5.53\% & 0.65\% \\
        \hline
        & & USCA & 39.80\% & 2.20\% \\
        & & Europe & 24.13\% & 1.33\% \\
        Market & 5.53\% & LATAM & 17.12\% & 0.95\% \\
        & & Africa & 11.74\% & 0.65\% \\
        & & Asia & 7.19\% & 0.40\% \\
        \hline
    \end{tabular}
    \caption{Weight distribution of evaluation dimensions and criteria derived from AHP-FCE method for Supplier Evaluation.}
    \label{tab:dimension_weights}
\end{table}

\subsubsection{Chain of Thought prompt}
Chain-of-Thought (CoT) prompting is a method of guiding a language model to reason step by step as it generates an answer, rather than directly providing the final answer~\cite{29-wei2022chain}. This approach is particularly suitable for problems requiring multi-step reasoning or complex computation. Using CoT prompting, the model is encouraged to simulate human thought processes, allowing it to break down complex problems and generate more accurate and logical answers.

\subsubsection{Fine-tuned Model with LoRA}
LoRA is an efficient fine-tuning method for large-scale pre-trained language models~\cite{24-han2024parameter}. As model sizes grow, traditional full-model fine-tuning becomes increasingly impractical due to the immense number of parameters that need adjustment, leading to extremely high computational and storage costs. LoRA addresses this challenge by enabling efficient model adaptation without the need to retrain all model parameters, significantly reducing resource requirements while maintaining performance.

\begin{table*}[!htb]
    \hspace*{-1cm} 
    \centering
    \begin{tabular}{l ccc ccc ccc}
        \toprule
         \multirow{2}{*}{Model} & \multicolumn{3}{c}{Supplier Evaluation} & \multicolumn{3}{c}{Customer Satisfaction } & \multicolumn{3}{c}{Air Quality} \\
        \cmidrule(lr){2-4} \cmidrule(lr){5-7} \cmidrule(lr){8-10}
        & Precision↑ & Recall↑ & F1↑ & Precision↑ & Recall↑ & F1↑ & Precision↑ & Recall↑ & F1↑ \\
        \midrule
        ChatGPT3.5-turbo &   0.589 & 0.593 & 0.555 &   0.538 & 0.441 & 0.306 & 0.521 & 0.540 & 0.515 \\
        Claude-3-sonnet & 0.521 &0.527 & 0.519 &0.624 &0.605 &0.601 & 0.660 & 0.539 & 0.501 \\
        ChatGPT3.5-turbo + CoT & 0.630 & 0.491 & 0.509 & 0.611  & 0.552 & 0.468  &  0.645 & 0.556 & 0.544 \\
        Claude-3-sonnet + CoT &0.632 &0.629 & 0.630 &0.715 & 0.653 &0.642 & 0.718 & 0.647 & 0.645 \\
        few-shot ChatGPT3.5-turbo &  0.629 & 0.602 & 0.609 & 0.566 & 0.610 & 0.521 & 0.687 & 0.700 & 0.686 \\
        few-shot Claude-3-sonnet &\textbf{0.750 }& \textbf{0.750} & \textbf{0.720} &\textbf{0.745} &\textbf{0.763} & \textbf{ 0.753}& 0.777 & 0.716 & 0.711 \\
         few-shot ChatGPT3.5-turbo + CoT &0.570 &0.484 &0.456& 0.549 & 0.602 & 0.504&0.709&0.696&0.697\\
        few-shot Claude-3-sonnet + CoT & 0.584 & 0.681 & 0.620 &0.641& 0.683 &0.645 &\textbf{0.804}&\textbf{0.775}&\textbf{0.777}\\
        
        \bottomrule
    \end{tabular}
    \caption{Performance comparison of API models.}
    \label{tab:combined_api}
\end{table*}

\section{EXPERIMENTS}

To fully validate our supplier evaluation methodology, we employed several major LLMs. For commercial APIs, we selected two representative models: ChatGPT-3.5-turbo and Claude-3-sonnet-20240229. For open source models, we used LLaMA3-8B~\cite{25-touvron2023llama}, Qwen2-7B~\cite{26-yang2024qwen2}, and ChatGLM4-9B~\cite{27-glm2024chatglm}. In addition, we developed LoRA-tuned versions of these open-source models to explore their optimization potential for different tasks. To systematically evaluate the performance of these models, we adopted several standard evaluation metrics, including precision, recall, and the F1 score. By comprehensively analyzing these metrics, we can thoroughly assess the performance differences between the LLM-based MCDM method and traditional expert evaluations. As shown in Table~\ref{tab:dimension_weights}, the AHP-FCE method is used to determine the weights of each dimension and criterion within each dimension. For supplier evaluation, as an example, the results are categorized into four criteria: Good, Fair, Average, and Poor. For the other two tested applications, the datasets include the final expert evaluation, which can be used as the groundtruth for comparison.

\subsection{Datasets}
First, in this part, we introduce the tested datasets. 

\textbf{Data Co-Supply Chain Dataset\footnote{https://www.kaggle.com/datasets/jolenechen/datacosupply\\chaindataset/data}:}  
The Data Co-Supply Chain Dataset provides multidimensional data, including suppliers' delivery timeliness, product quality, and supply capability. This data set enables the analysis of supplier reliability and performance, offering decision support for supply chain management. It ensures the selection of suppliers that consistently deliver high-quality products and good customer service, ultimately optimizing supply chain efficiency. 

\textbf{Customer Feedback and Satisfaction Dataset\footnote{https://www.kaggle.com/datasets/jahnavipaliwal/customer-feedback-and-satisfaction}:}
The Customer Feedback and Satisfaction Dataset is a synthetic data set designed to analyze and predict customer satisfaction, providing data scientists with insights into customer behavior and data-driven support to formulate business strategies. 

\textbf{Air Quality and Pollution Evaluation\footnote{https://www.kaggle.com/datasets/mujtabamatin/air-quality-and-pollution-assessment/data}:} 
Air Quality and Pollution Evaluation involves a comprehensive analysis to reduce sources of pollution, improve air quality and protect public health and environmental sustainability. This is achieved by monitoring and analyzing the distribution of various air pollutants, population densities, and industrial areas.

\subsection{API Models Evaluation}
 As shown in Table~\ref{tab:combined_api}, the performance of different API models on Supplier Evaluation, Customer Satisfaction Evaluation, and Air Quality Evaluation was compared using evaluation metrics such as Precision, Recall, and F1 score. Few-shot Claude-3-sonnet achieved the highest scores in all three metrics in both the Supplier Evaluation and Customer Satisfaction Evaluation tasks. In the Air Quality Evaluation task, the few-shot Claude-3-sonnet + CoT performed the best on all three metrics. 
 
 \subsection{Open-Source Models Evaluation}
\begin{figure}[!b]
    \centering
    \includegraphics[width=0.49\textwidth]{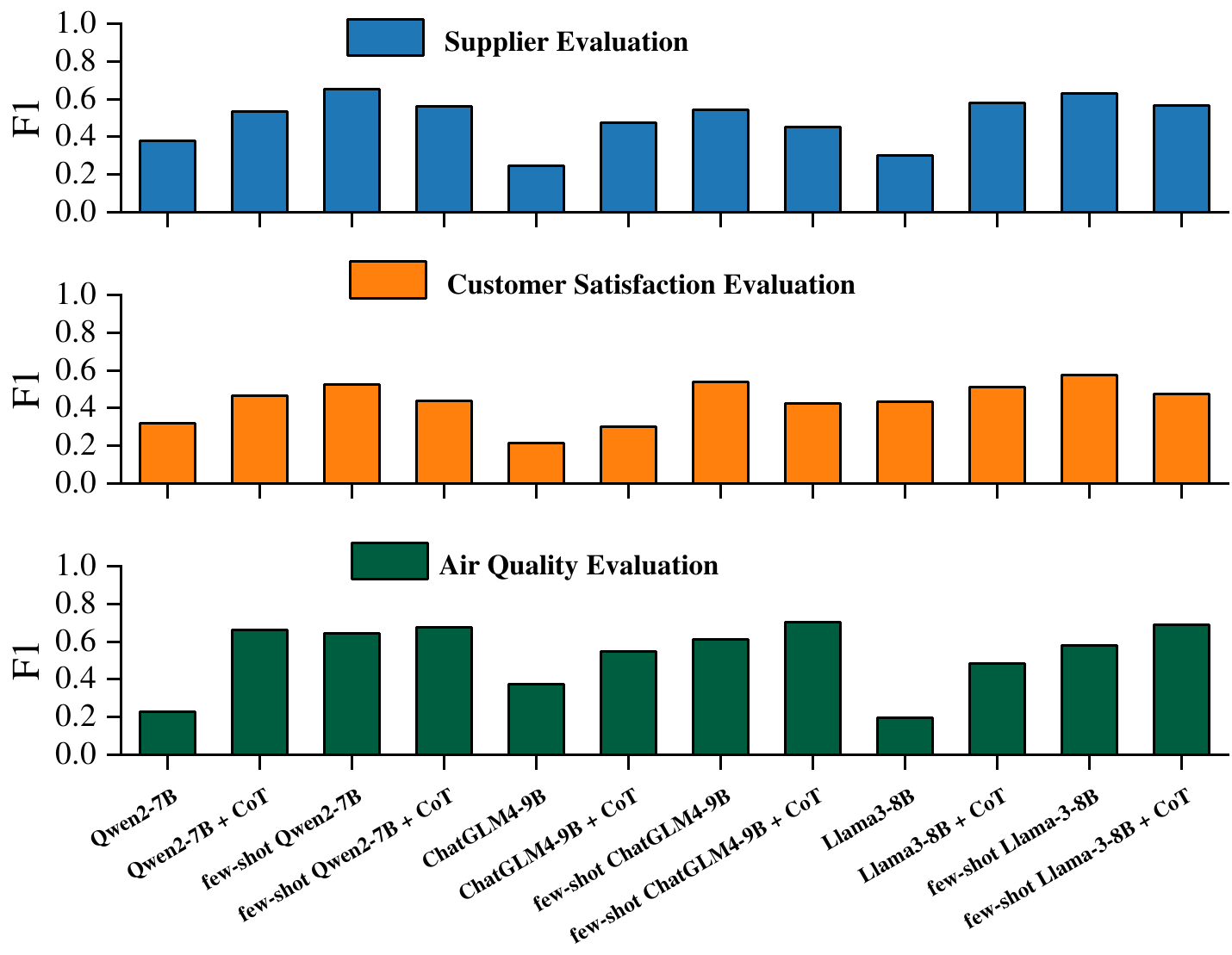} 
    \caption{Performance comparison of different open-source models in three datasets.}
    \label{fig:2}
\end{figure}

\begin{figure*}[!bth]
    \centering
    \begin{minipage}{0.32\textwidth}
        \centering
        \includegraphics[width=\linewidth]{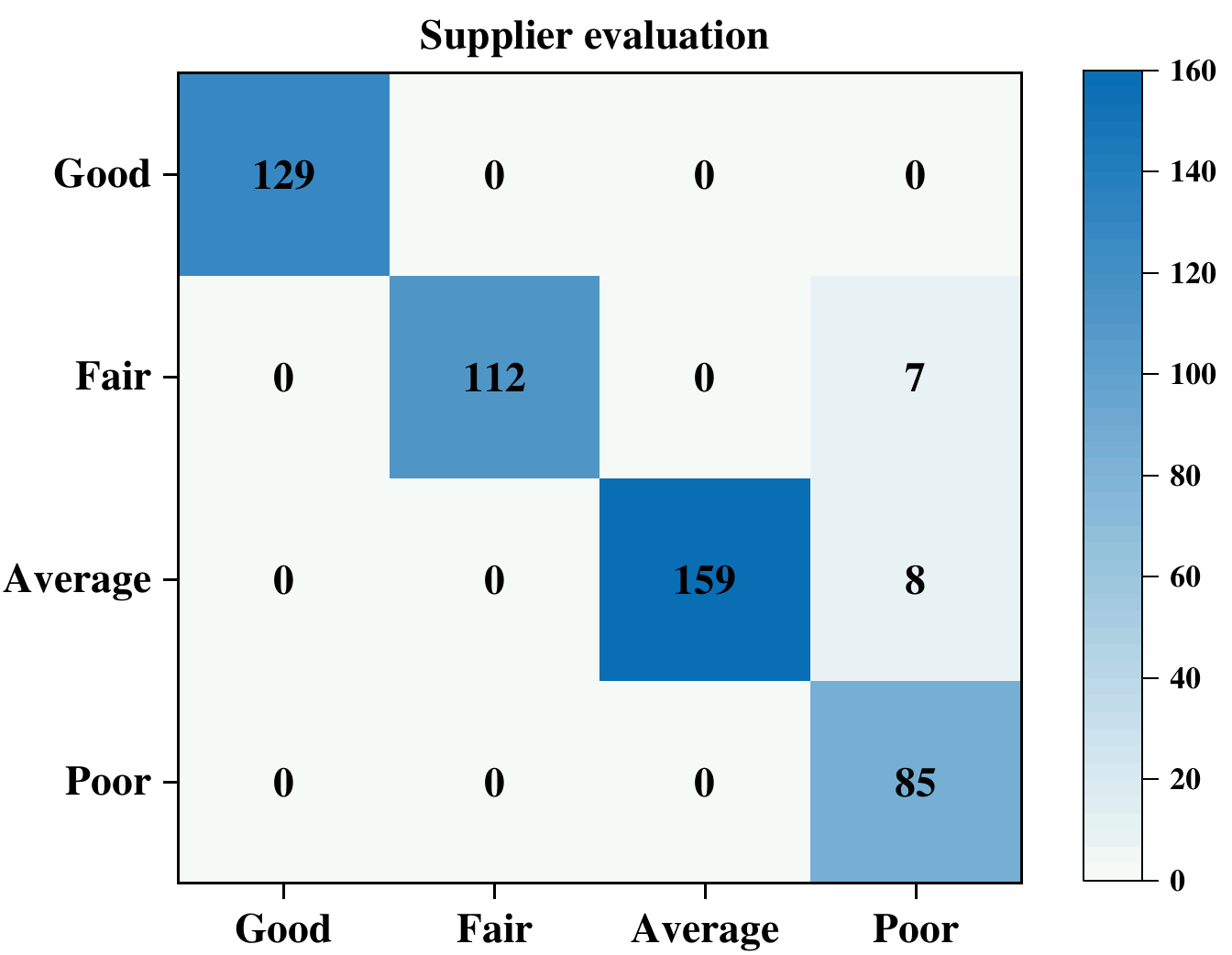}
        \subcaption{LoRA Qwen2-7B} \label{fig:sub1}
    \end{minipage}%
    \hfill
    \begin{minipage}{0.32\textwidth}
        \centering
        \includegraphics[width=\linewidth]{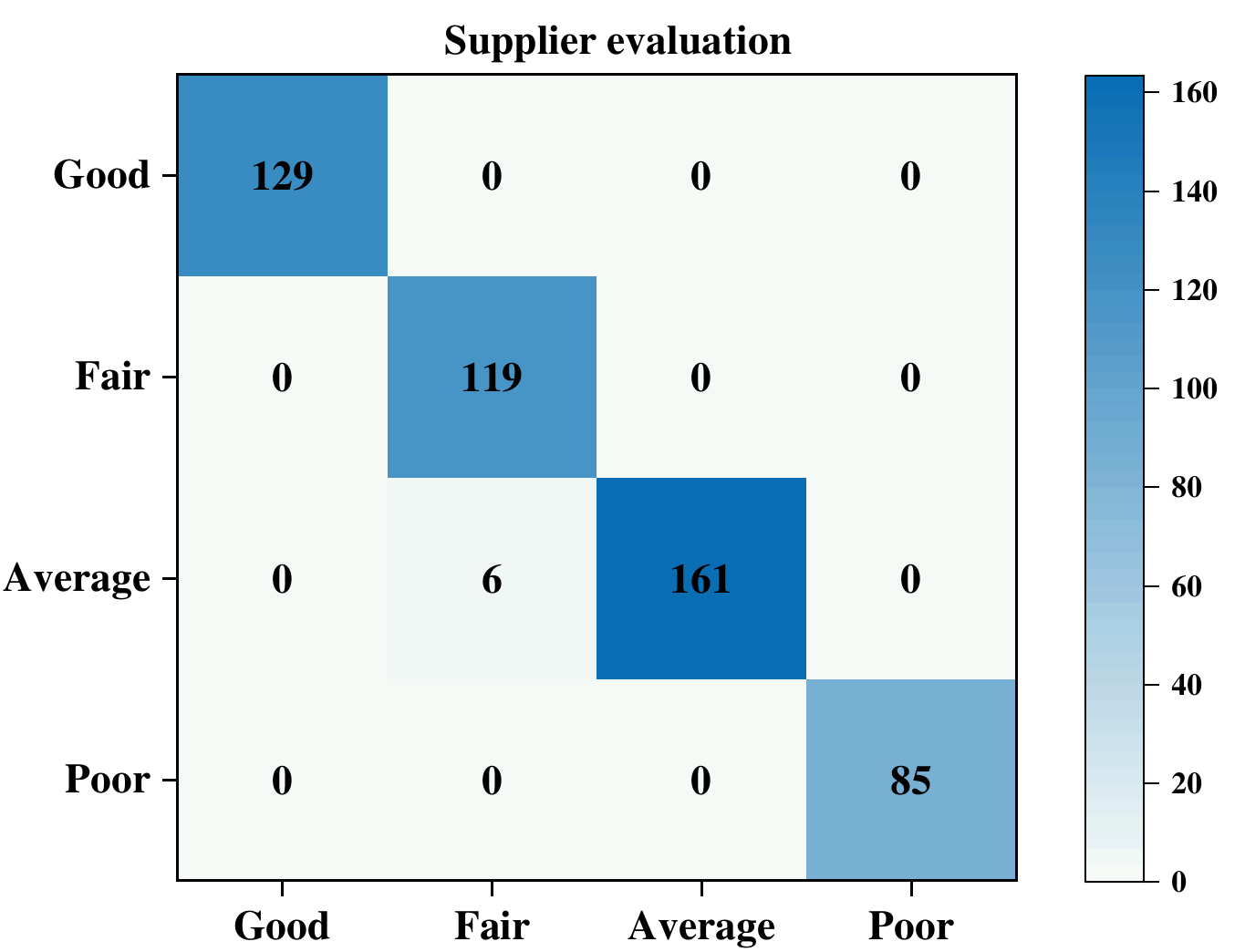}
        \subcaption{LoRA ChatGLM4-9B} \label{fig:sub2}
    \end{minipage}%
    \hfill
    \begin{minipage}{0.32\textwidth}
        \centering
        \includegraphics[width=\linewidth]{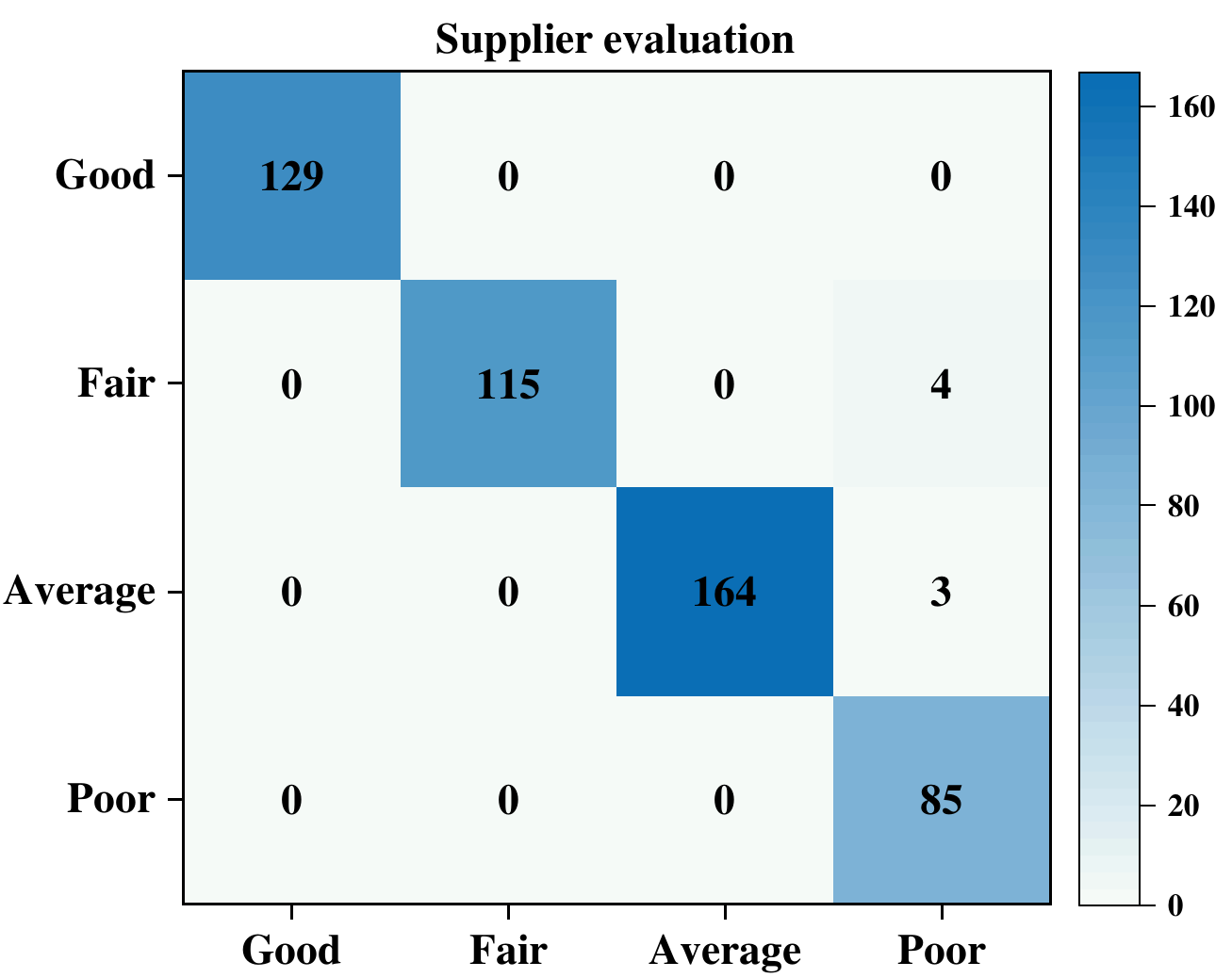}
        \subcaption{LoRA Llama3-8B} \label{fig:sub3}
    \end{minipage}
    
    \vspace{0.5cm}

    \begin{minipage}{0.32\textwidth}
        \centering
        \includegraphics[width=\linewidth]{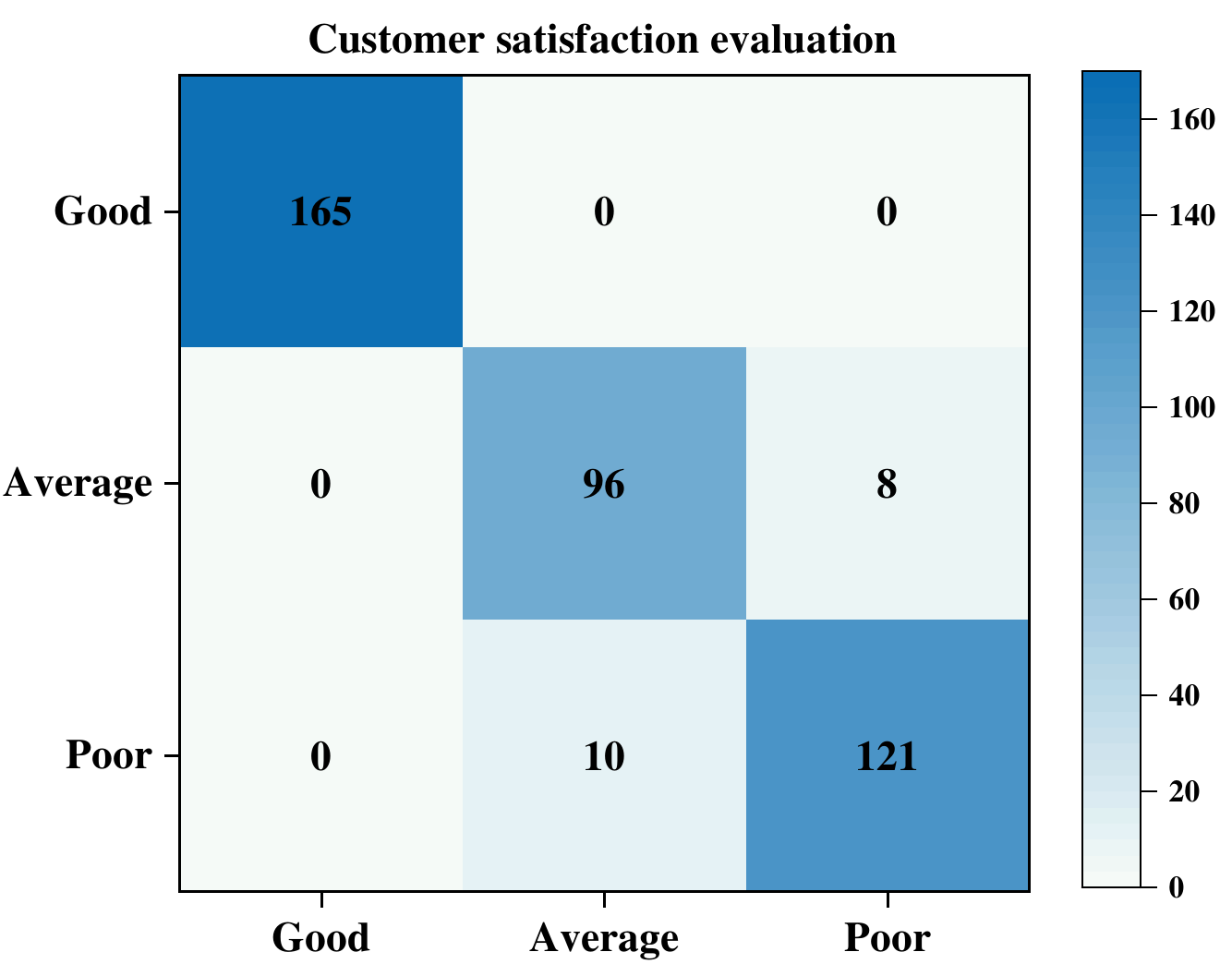}
        \subcaption{LoRA Qwen2-7B} \label{fig:sub4}
    \end{minipage}%
    \hfill
    \begin{minipage}{0.32\textwidth}
        \centering
        \includegraphics[width=\linewidth]{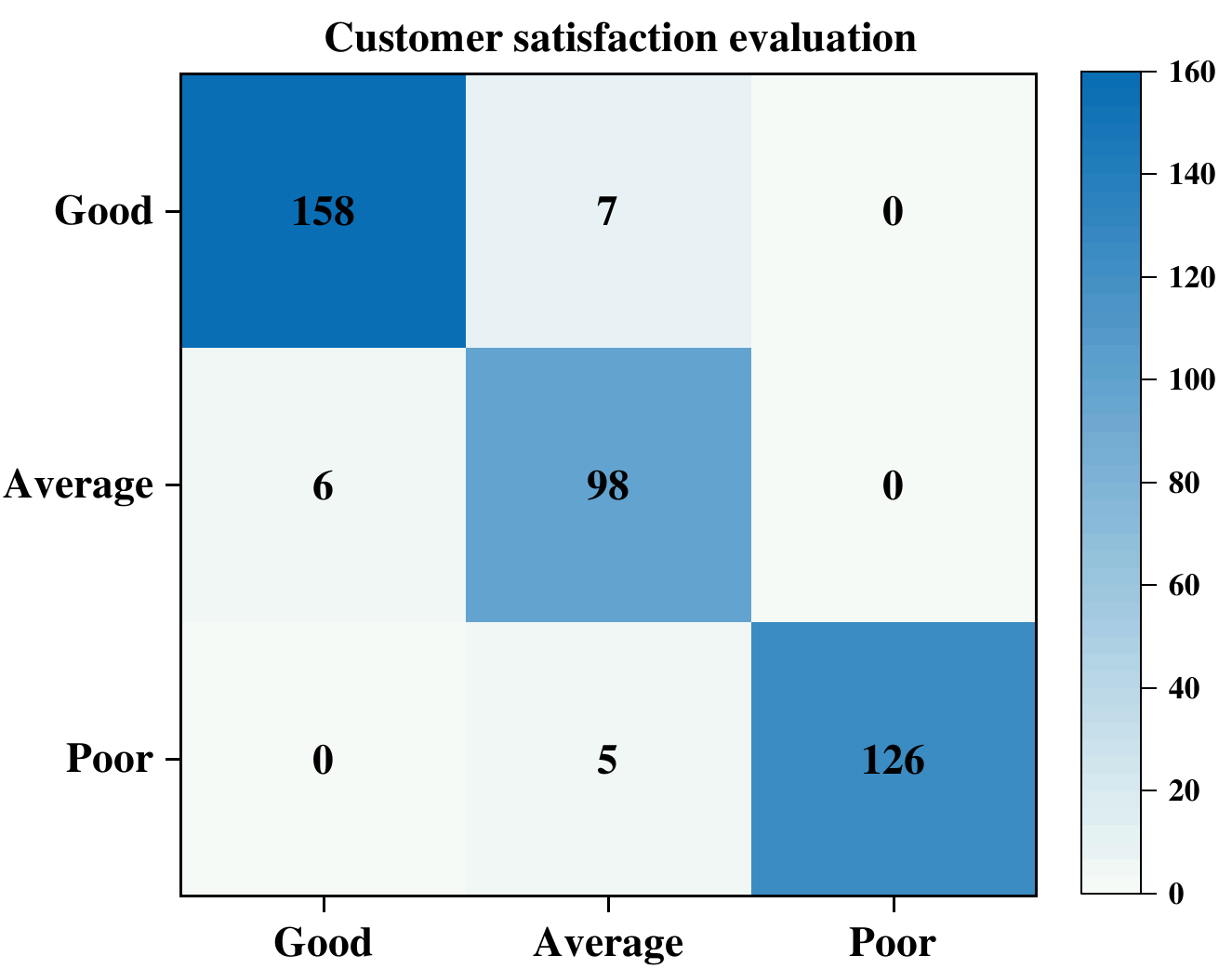}
        \subcaption{LoRA ChatGLM4-9B} \label{fig:sub5}
    \end{minipage}%
    \hfill
    \begin{minipage}{0.32\textwidth}
        \centering
        \includegraphics[width=\linewidth]{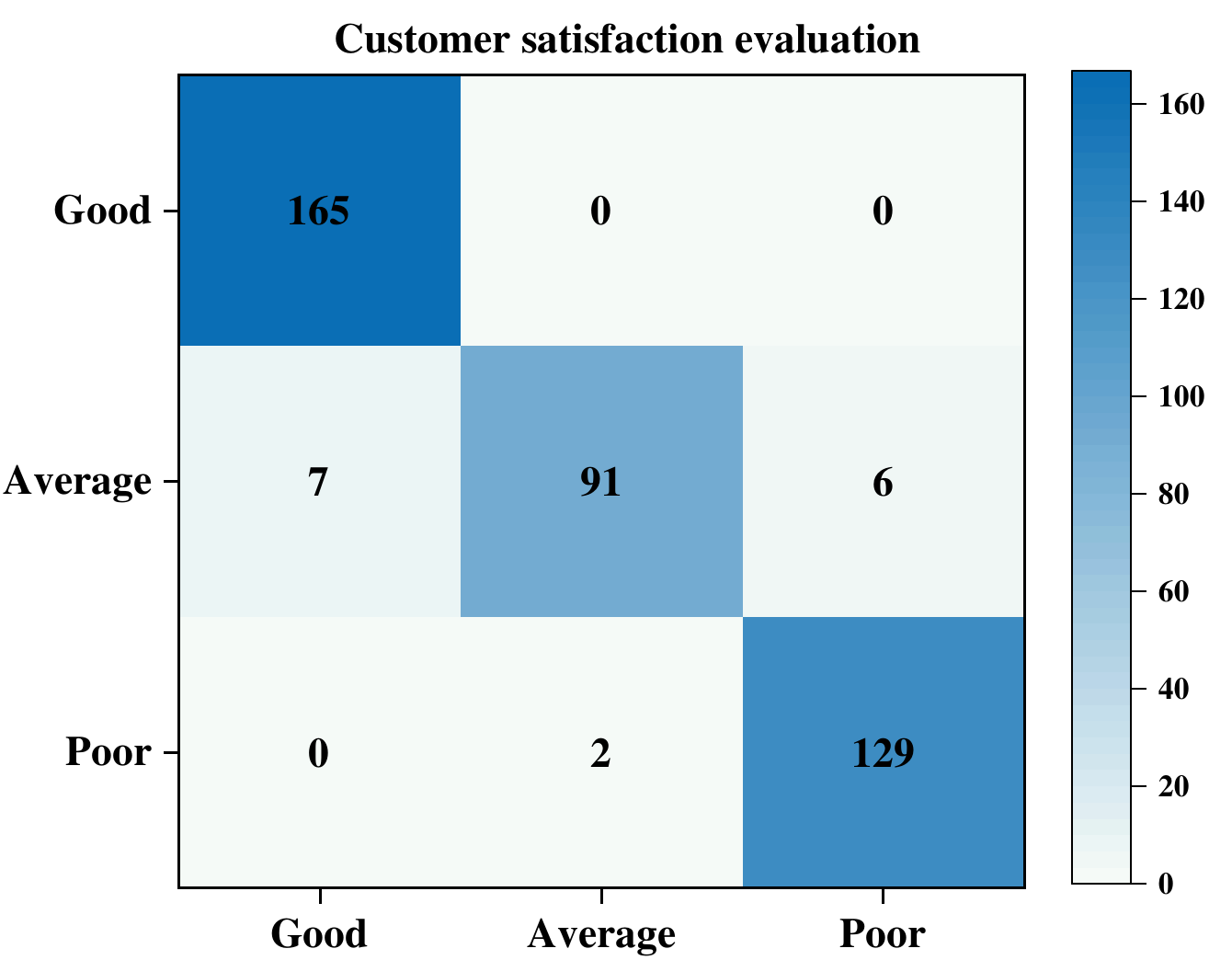}
        \subcaption{LoRA Llama3-8B} \label{fig:sub6}
    \end{minipage}
    
    \vspace{0.5cm}

    \begin{minipage}{0.32\textwidth}
        \centering
        \includegraphics[width=\linewidth]{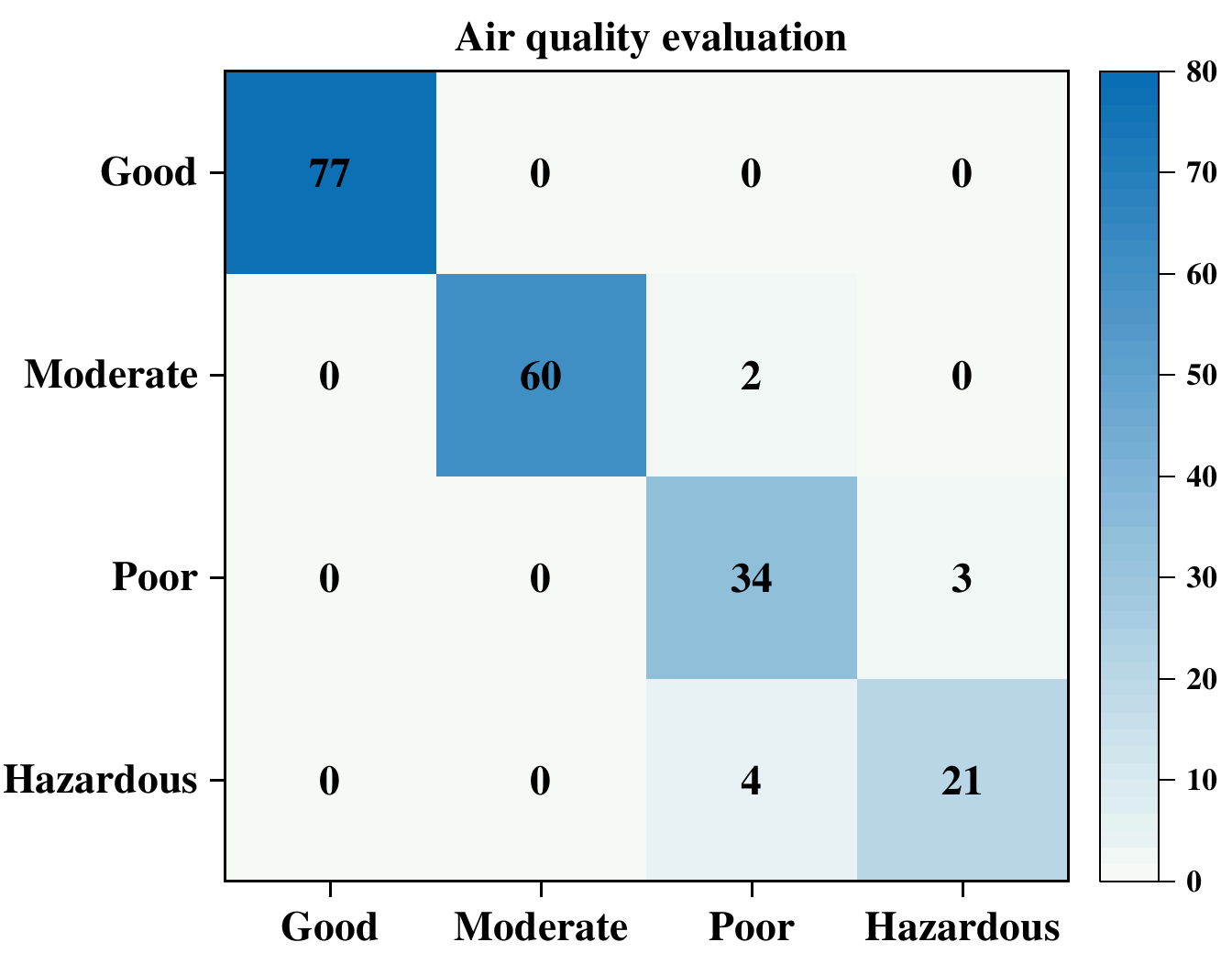}
        \subcaption{LoRA Qwen2-7B} \label{fig:sub7}
    \end{minipage}%
    \hfill
    \begin{minipage}{0.32\textwidth}
        \centering
        \includegraphics[width=\linewidth]{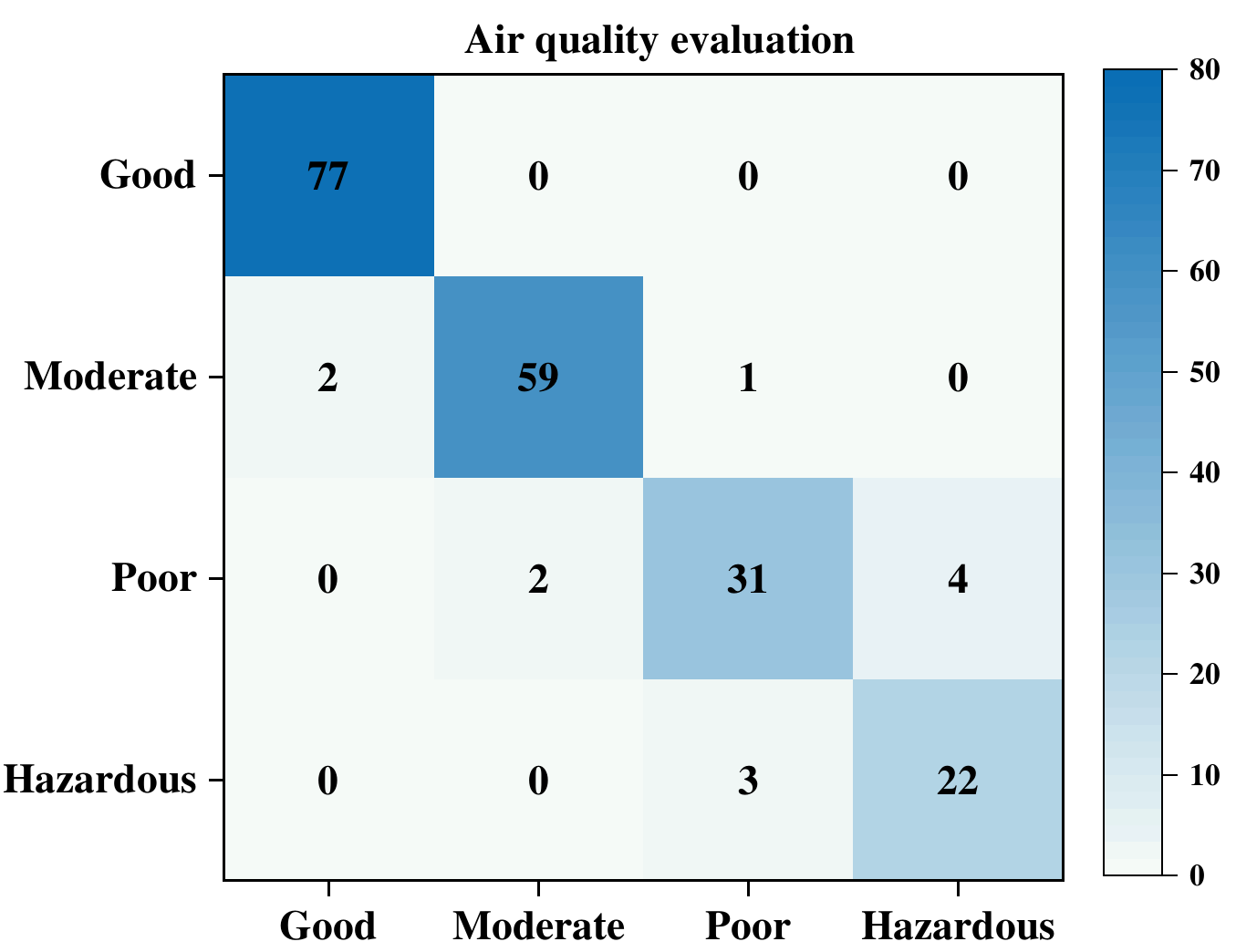}
        \subcaption{LoRA ChatGLM4-9B} \label{fig:sub8}
    \end{minipage}%
    \hfill
    \begin{minipage}{0.32\textwidth}
        \centering
        \includegraphics[width=\linewidth]{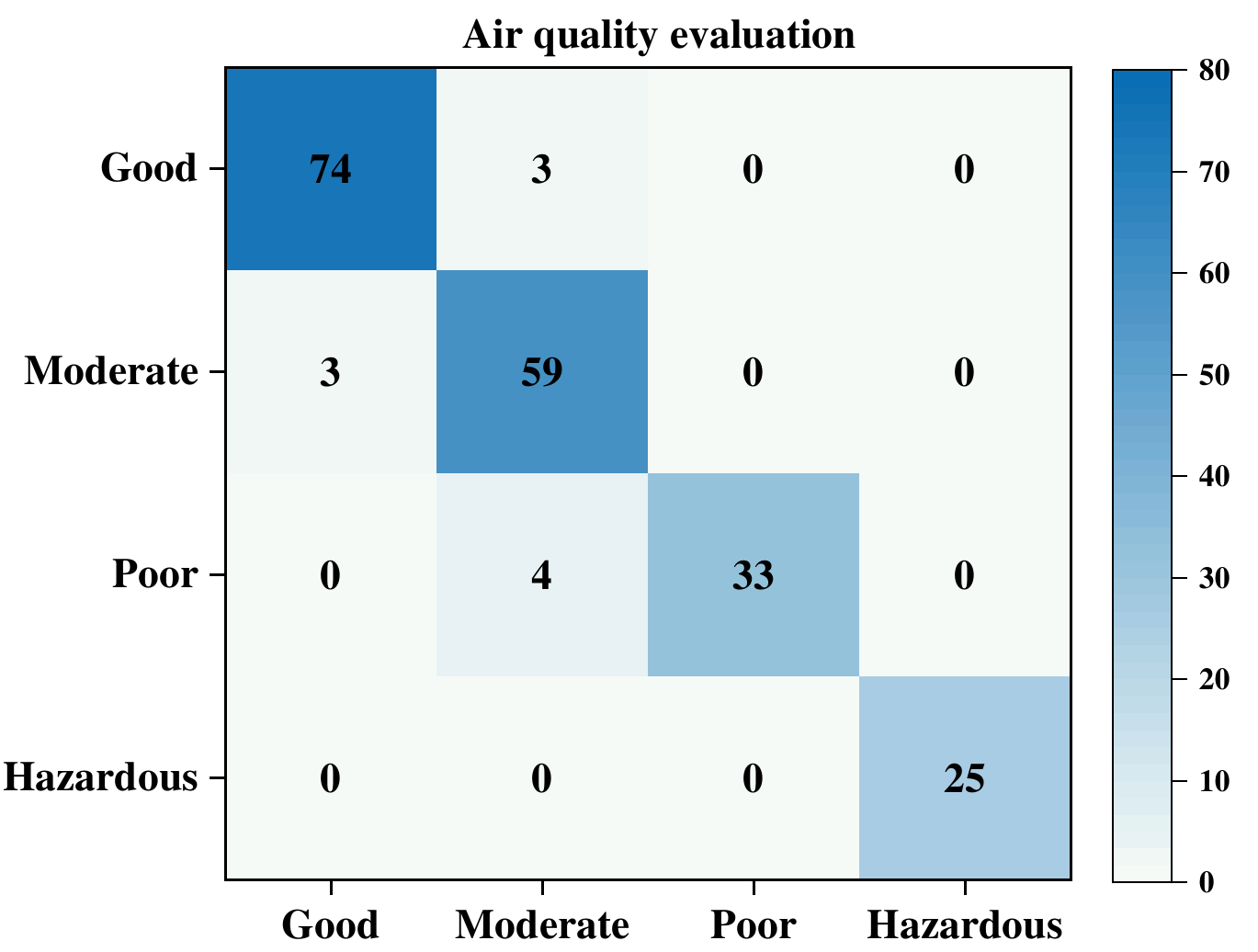}
        \subcaption{LoRA Llama3-8B} \label{fig:sub9}
    \end{minipage}
    \caption{Comparison of the confusion matrices of different fine-tuning models on the three datasets.}
    \label{fig:3}
\end{figure*}

The superior performance of Claude-3-sonnet + CoT can be attributed to the integration of the CoT inference enhancement method, which enhances its reasoning capabilities. However, the primary factor driving the model's exceptional performance lies in the few-shot learning approach. Fine-tuning the model with a small number of examples significantly improved its adaptability and accuracy across all tasks. In particular, Claude-3-sonnet with few shots achieved the highest metrics in supplier evaluation and customer satisfaction assessment, showcasing its ability to effectively learn task-specific patterns even with limited training data. When combined with CoT, the few-shot Claude-3-sonnet + CoT demonstrated further improvements, particularly in the air quality evaluation task, where it achieved the highest scores in precision, recall and F1 score. Although CoT provided additional reasoning support, the foundation of this performance was established through the fine-tuning process. This highlights that few-shot learning played a critical role in enabling the model to process complex decision-making scenarios with improved accuracy and efficiency.

However, we see that the F1 scores are around 55\% for pure APT models, and the F1 scores rise to around 75\% while employing few-shot and CoT prompts. Indicting that the performance of basic LLMs in solving MCDM is not as good as experts, and different prompt methods on different models achieve significantly different performance.

\begin{table*}[!t]
    \centering
    \begin{tabular}{l ccc ccc ccc}
        \toprule
         \multirow{2}{*}{Fine-tuned Model} & \multicolumn{3}{c}{Supplier Evaluation} & \multicolumn{3}{c}{Customer Satisfaction} & \multicolumn{3}{c}{Air Quality} \\
        \cmidrule(lr){2-4} \cmidrule(lr){5-7} \cmidrule(lr){8-10}
        & Precision↑ & Recall↑ & F1↑ & Precision↑ & Recall↑ & F1↑ & Precision↑ & Recall↑ & F1↑ \\
        \midrule
        Qwen2-7B &   0.974 & 0.970 & 0.972 &   0.948 & 0.949 & 0.948 & 0.931 & 0.932 & 0.931 \\
        ChatGLM4-9B & 0.989 &0.988 &\textbf{ 0.988} &  0.951 &0.954 &0.952 & 0.918 & 0.917 &  0.918 \\
        Llama3-8B & 0.987 & 0.986 & 0.986 & 0.965  & 0.953 &\textbf{ 0.958}  &  0.964 & 0.951 & \textbf{0.956} \\
        \bottomrule
    \end{tabular}
    \caption{Performance comparison of LoRA fine-tuned models.}
    \label{tab:LoRA}
\end{table*}

In Fig.~\ref{fig:2}, we analyze three open-source models LLaMA3-8B, Qwen2-7B, and ChatGLM4-9B using the same prompt methods applied to API models. As shown in Table~\ref{tab:combined_api}, the size of the model does have a significant impact on its performance. Both API models and open-source models experience significant improvements using CoT and few-shot learning. Specifically, air quality evaluation performance increased from 0.228 to 0.704, highlighting the potential of models to process multidimensional data more effectively when coupled with the right learning strategies. We see both API and open-source models exhibit the same general performance trend with the integration of CoT and few-shot, the outcomes vary across different tasks. This consistency across model types offers robust support and a valuable reference for future applications in other MCDM tasks. 

However, we see that the F1 scores achieved by open-source models are less than the corresponding results achieved by API models, which is expected since the commercial APIs we employed are known to be more intelligent than those open-source models. The consistent results further indicate that the performance of basic LLMs~(both APIs and open-source models) in solving MCDM is not as good as that of experts, and different prompt methods on different models achieve significantly different performance. Therefore, it is urgent to design models that can perform human-expert level ability in solving MCDM tasks. 

\begin{figure}[!b]
    \centering
    \includegraphics[width=\linewidth]{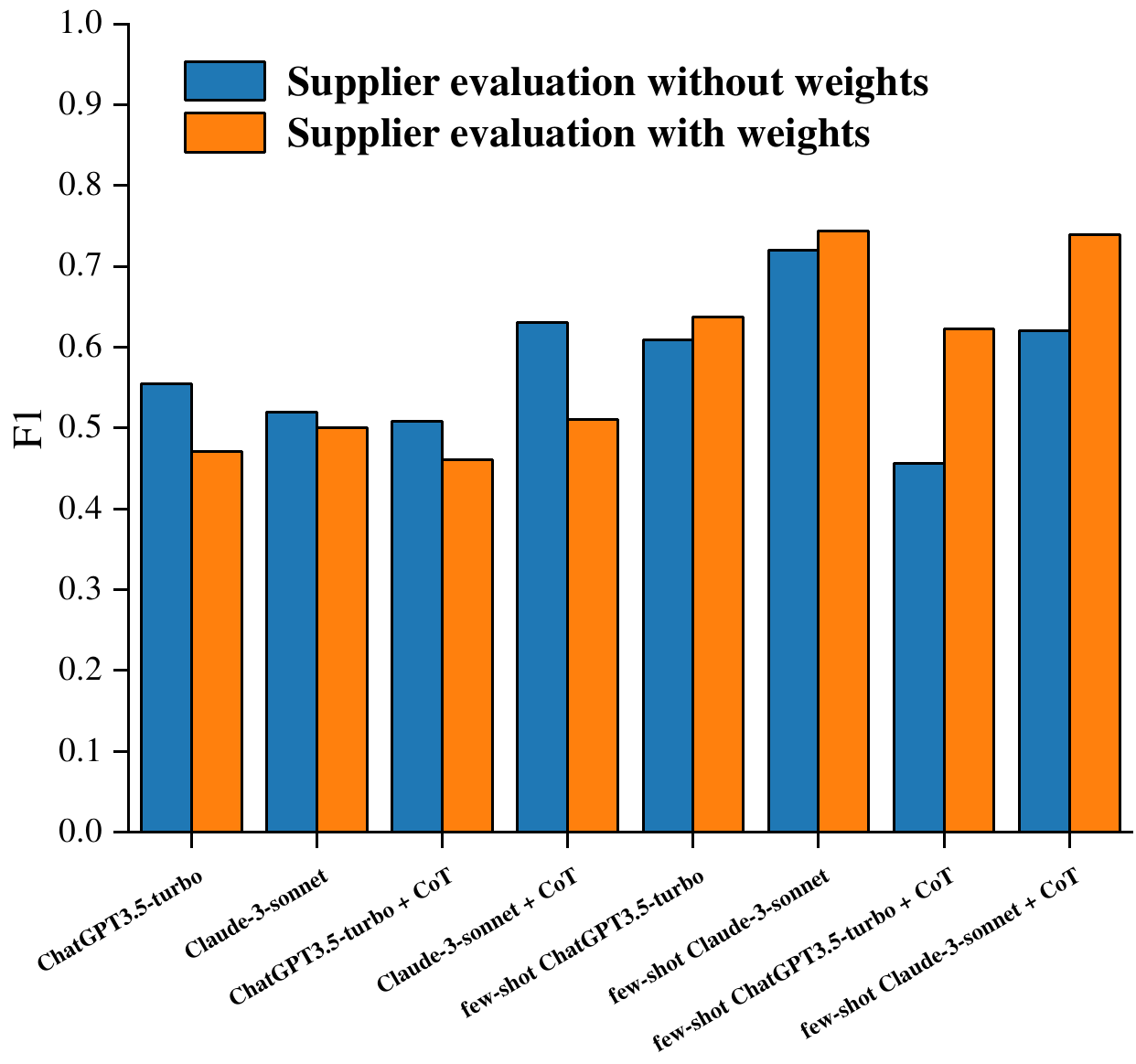} 
    \caption{Performance comparisons between weighted and unweighted cues in the supplier evaluation dataset.}
    \label{fig:4}
\end{figure}

In this paper, we employ the LoRA fine-tuning method to train open-source models. In the experiments, each model was fine-tuned using 500 examples from the three datasets, specifically targeting the enhancement of performance for each task. The results in Table~\ref{tab:LoRA} reveal that by integrating a few examples into the LoRA tuning process, the model can achieve an accuracy rate close to 0.99 for the supplier evaluation task, while maintaining scores above 0.9 for the other two datasets. This outcome indicates that through LoRA fine-tuning, the model can reach human-expert-level performance in complex MCDM tasks.

In addition, Fig.~\ref{fig:3} shows the confusion matrix for three different LoRA fine-tuned models on three different datasets. Each subfigure specifically demonstrates the performance of the model on a particular dataset, and the confusion matrices detail the relationship between the true labels and the model's predicted labels for each category. For further clarification, each confusion matrix not only reveals the accuracy and error of the model in a particular task, but also provides key performance metrics such as precision, recall, and F1 score.  In addition, by comparing the confusion matrices of different models on the same dataset, we visualized the strengths and weaknesses of each model in recognizing each category.

Furthermore, in order to show the importance of the weights on each criterion in the proposed framework, we compared the effectiveness of weighted and unweighted prompts in the supplier evaluation task in Fig.~\ref{fig:4}. These weights were configured according to the data provided in Table~\ref{tab:dimension_weights}, which analyzed the different dimensions of supplier evaluation and their relative importance. In our experiments, we tested the performance of both weighted and unweighted prompts before and after applying the few-shot learning method. The results indicate that, while both types of prompt showed improvement with few-shot learning, weighted prompts demonstrated a more significant performance boost. with well-designed prompt strategies to improve the model's ability to handle MCDM tasks.

\section{Conclusions}
This paper proposed a human-expert-level framework to deal with MCDM tasks through proper prompt design and fine-tuning with a small amount of data. The performance of LLMs has improved significantly from around 55\% to 75\% and finally stabilized around 95\%, achieving impressive evaluation performance. These findings demonstrate that LLMs can not only address the limitations of traditional MCDM methods in handling complex multidimensional tasks, but also highlight its great potential and advantages in decision analysis. Traditional MCDM methods, such as AHP and FCE, although widely used, are heavily based on manually set weights and expert experience. In this paper, three pre-trained models are optimized using fine-tuning techniques, and the results show performance comparable to human expert evaluators. This paper not only provides strong empirical evidence for the application of LLMs in MCDM but also offers valuable insights for the development of intelligent decision-making systems with LLMs.

For future work, applying the proposed framework to more real applications is promising. In addition, it is also interesting to generate and tag fine-tuning MCDM data on LLMs, since although fine-tuning does not need much data, for some applications, collecting and tagging real data is costly.

\bibliographystyle{named}
\bibliography{ijcai24}

\newpage
\onecolumn 
\section{Appendix}

Figure 5 demonstrates examples of air quality evaluation from three datasets, utilizing Zero-shot, Few-shot, and Chain of Thought methods. Each example evaluates air quality based on input environmental data and assigns a final air quality rating (Good, Moderate, Poor, or Hazardous). The Zero-shot method directly outputs the rating based on the input data, the Few-shot method references similar examples before evaluation, and the Chain of Thought method analyzes the impact of each environmental factor step-by-step to derive the final rating. These examples illustrate the application and outcomes of different methods in air quality evaluation.
\begin{figure*}[ht]
    \centering
    \includegraphics[width=\textwidth]{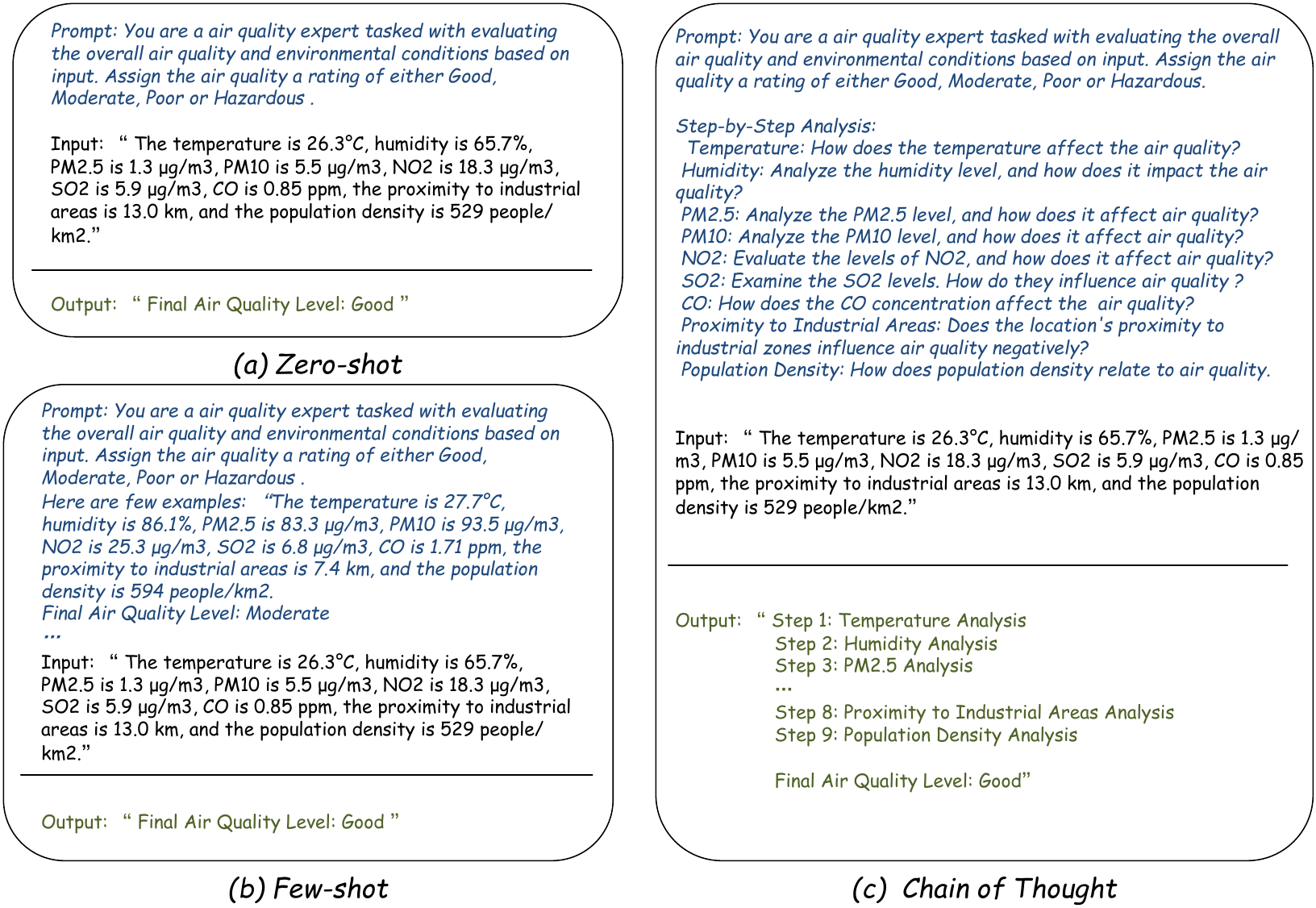} 
    \caption{Evaluation methods: Zero-shot, Few-shot, and Chain of Thought.}
    \label{fig:5}
\end{figure*}

\newpage
In data analysis in the field of supply chain management, two different methods of prompts are used for supplier evaluation. The first method is prompts with weights, where different weights are assigned based on dimensions and specific criteria to highlight the importance of certain factors. This approach allows optimizing the model's response to a specific problem by precisely adjusting the weights. Another approach is to use generic prompts without templates and to look at supplier performance holistically without preset weights. This approach shows its strengths when dealing with complex issues that require a broad perspective and flexibility in response. The results of validating the model with prompts that require refinement are shown in Fig.~\ref{fig:4} to demonstrate whether the results obtained are better than the performance without weights.

\begin{figure}[htb!]
    \centering
    \includegraphics[width=\linewidth]{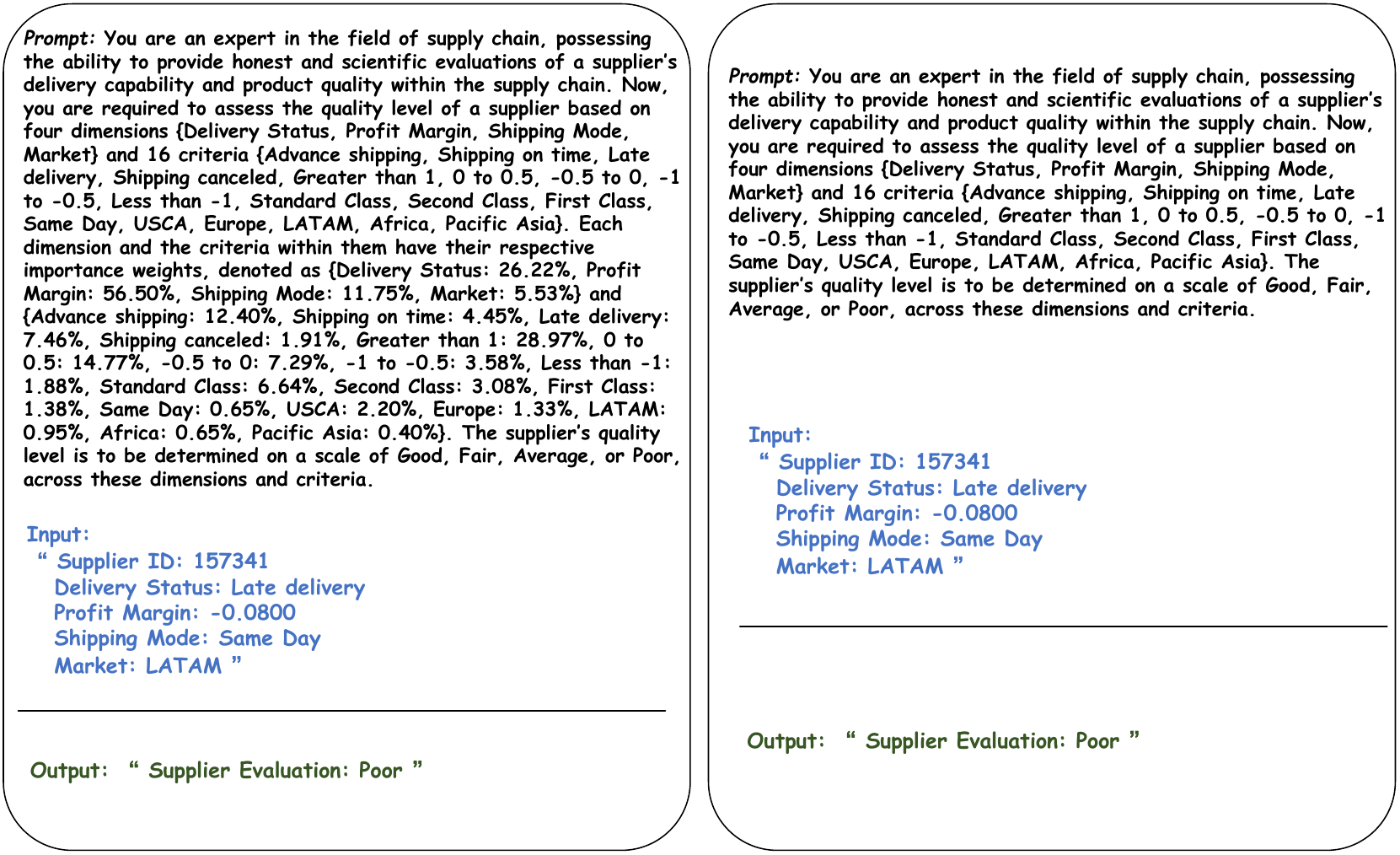} 
    \caption{Two different prompts, one with weights and one without.}
    \label{fig:7}
\end{figure}

\subsubsection{datasets}
\noindent\textbf{Data Co-Supply Chain Dataset:} \\
Data available at: \href{https://www.kaggle.com/datasets/jolenechen/datacosupplychaindataset/data}{https://www.kaggle.com/datasets/jolenechen/datacosupplychaindataset/data}

\noindent\textbf{Customer Feedback and Satisfaction Dataset:} \\
Data available at: \href{https://www.kaggle.com/datasets/jahnavipaliwal/customer-feedback-and-satisfaction}{https://www.kaggle.com/datasets/jahnavipaliwal/customer-feedback-and-satisfaction}

\noindent\textbf{Air Quality and Pollution Evaluation:} \\
Data available at: \href{https://www.kaggle.com/datasets/mujtabamatin/air-quality-and-pollution-assessment/data}{https://www.kaggle.com/datasets/mujtabamatin/air-quality-and-pollution-assessment/data}

\end{document}